# The Landmark Selection Method for Multiple Output Prediction

**Krishnakumar Balasubramanian** KRISHNAKUMAR3@GATECH.EDU
**Guy Lebanon** LEBANON@CC.GATECH.EDU
College of Computing, Georgia Institute of Technology, Atlanta, Georgia, USA

## Abstract

Conditional modeling $x \mapsto y$ is a central problem in machine learning. A substantial research effort is devoted to such modeling when $x$ is high dimensional. We consider, instead, the case of a high dimensional $y$, where $x$ is either low dimensional or high dimensional. Our approach is based on selecting a small subset $y_L$ of the dimensions of $y$, and proceed by modeling (i) $x \mapsto y_L$ and (ii) $y_L \mapsto y$. Composing these two models, we obtain a conditional model $x \mapsto y$ that possesses convenient statistical properties. Multi-label classification and multivariate regression experiments on several datasets show that this method outperforms the one vs. all approach as well as several sophisticated multiple output prediction methods.

## 1. Introduction

Conditional modeling $x \mapsto y$ is a central problem in machine learning. Specific cases include classification, where $y$ is a discrete random variable, and regression, where $y$ is a continuous random variable. Much of the attention in recent years has focused on the case where $x$ is a high dimensional vector. In this case, traditional statistical methods are inefficient due to overfitting. Proposed alternatives for high dimensional $x$ include feature selection and regularized models.

We consider, instead, the case of a high dimensional $y$, where $x$ is either low dimensional or high dimensional. The baseline approach in this case is to independently construct models $x \mapsto y_i \in \mathbb{R}$ for $i = 1, \ldots, k$ (assuming $y$ is a $k$-dimensional real vector). This approach has the advantage of drawing from a wide variety of available single output models, including linear and non-linear regression, logistic regression, and support vector machines. The main disadvantage is that the independent models do not take advantage of a likely correlation between the dimensions of $y$. Incorporating this correlation becomes especially important when the dimensionality of $y$ is higher or of similar order to the dimensionality of $x$.

Our approach is based on selecting a small subset $L \subset \{1, \ldots, k\}$ of the dimensions of $y$, and constructing two models:

$$x \mapsto y_L \tag{1}$$
$$y_L \mapsto y, \tag{2}$$

where we use the standard notation $y_L = \{y_i : i \in L\}$. We thus have three problems: selecting the subset $L$, estimating (1), and estimating (2).

Specifically, we estimate model (2) in conjunction with selecting $L$ via least-squares regression with group Lasso based hierarchical regularization. The precise model (1) varies, based on whether $y$ is discrete or continuous. It may be any low-dimensional multiple output model, such as multilabel logistic regression and SVM, or multiple linear regression. If the dimensionality of $x$ is high, regularization for model (1) is also necessary.

The underlying assumption of our model is that there exists a subset $L$ of the dimensions of $y$, called landmark variables, such that the remaining dimensions of $y$ may be expressed as a noisy linear combination $y = Ay_L + \epsilon$, with sparse coefficients. Several practical data sets exhibit such a kind of relationship. One example is the prediction of future stock prices $y$ from current stock prices $x$. The relationship $y = Ay_L + \epsilon$ is motivated by the identical trends of stock prices of multiple companies with a similar business model, or of multiple investment banks with similar holding portfolio. This phenomenon has been well documented in finance under the term cointegration. Another example is the classification of images ($x$) depending on what objects appear or do not appear in them ($y$). Obviously, some objects tend to appear or to not appear simultaneously, such as sky and tree, or car and road.

The cardinality $s$ of the subset $L$ is typically orders of magnitude lesser than the actual dimension of the output space making the method scale well to ultra-high dimensional outputs. For example, the naive one vs. all method requires $O(k)$ independent models that need to be learnt from the data, whereas the number of subproblems selected in the proposed approach scales at the rate of $O(s)$. Assuming $s \ll d$ we see that there is a huge advantage in terms of number of subproblems selected.

We report in this paper experimental results for classification and regression on multiple datasets. Based on our experimental study, we conclude that our model outperforms the one vs. all approach as well as several sophisticated multiple output prediction methods.

## 2. Related Work

Several methods have been proposed for multi-output prediction both in regression and classification setting. In the regression setting, most approaches have focused on penalization of the regression matrix or input space sharing. For example (Izenman, 1975) introduced low-rank penalization of the regression matrix, which was analyzed in (Reinsel & Velu, 1998) in the low-dimensional setting. Recent work focused on analyzing penalized regression in high dimensions (Rohde & Tsybakov, 2011). An alternative approach that is directly applicable to multi-output prediction is group lasso (Yuan & Lin, 2006). Though these methods are popular and widely applicable, they do not directly model correlations in the output dimensions, which can be used to reduce the complexity of the problem. A notable exception is the curds and whey method (Breiman & Friedman, 1997) which uses shrinkage techniques in output space to reduce prediction error.

In the classification setting, the popular approach of one-vs-all was proposed by several researchers (see (Rifkin & Klautau, 2004) for a discussion). This method ignores the dependencies between the different dimensions of $y$, and is inefficient when $y$ is high dimensional. A summary of improvements over the one-versus-all method is available in (Tsoumakas et al., 2010). Alternative approaches assume a class hierarchy on the output space (Cesa-Bianchi et al., 2006), graph structure on the output space (Vens et al., 2008) and joint feature extraction from output and input spaces in large margin setting (Tsochantaridis et al., 2006).

A paper related to our proposed method is (Hsu et al., 2009), which consider multi-label prediction in a sparse high-dimensional output space. Their proposed method for multi-label classification is to randomly project $y$ and construct a regression model on the reduced subspace. There are two significant differences between this paper and our approach: (i) our approach uses data-dependent transformation, rather than a random projection, and (ii) our approach selects a subset of the dimensions of $y$ that contributes to computational efficiency, statistical analysis, and is in line with some practical scenarios (see previous section). Furthermore, the approach by (Hsu et al., 2009) might not be applicable in the regression setting, as output sparsity assumption does not hold for regression in practice.

Recently proposed variations on (Hsu et al., 2009) include (Tai & Lin, 2012) that propose to reduce the dimensionality of the output space by PCA, and (Bi & Kwok, 2011) that propose to reduce the label space by preserving a graph structure hierarchy on $y$. While these methods are sub-linear, they still project on to a low-dimensional real subspace, and hence they do not guarantee that the problem in the reduced subspace is easier than the original problem.

Our approach also has a close connection to sparse PCA (Zou et al., 2006). Two significant differences are: (i) sparse PCA is generally applied to the covariates $x$ rather than $y$, and (ii) our focus is on identifying the landmarks $L$ and the relationship $y_L \mapsto y$ rather than estimating the principal components themselves.

## 3. The landmark selection method:

We consider a common scenario where $x \in \mathbb{R}^d$ and $y \in \mathcal{Y}$, where $\mathcal{Y}$ is either $\mathbb{R}^k$ (regression) or $\{0,1\}^k$ (classification), and $k, d$ are high dimensional. We denote the data matrices, containing $n$ labeled samples, by $X \in \mathbb{R}^{n \times d}$ and $Y \in \mathbb{R}^{n \times k}$.

### Step 1: Selecting the landmark set $L$ and modeling (2)

A convenient way to select the set of landmark dimensions $L$, and to model (2) simultaneously is the following regularized least squares regression model

$$\hat{A} = \arg\min_{A \in \mathbb{R}^{k \times k}} \|Y - YA\|_F^2 + \lambda_1 \|A\|_{1,2} + \lambda_2 \|A\|_1 \quad (3)$$

where

$$\|A\|_F \stackrel{\text{def}}{=} \sqrt{\mathsf{tr}(A^\top A)}$$

$$\|A\|_{1,2} \stackrel{\text{def}}{=} \sum_{i=1}^{k} \sqrt{\sum_{j=1}^{k} A_{i,j}^2}$$

$$\|A\|_1 \stackrel{\text{def}}{=} \sum_{i=1}^{k} \sum_{j=1}^{k} |A_{ij}|.$$

The first term in (3) is the least squares empirical risk that is standard in linear regression models. Obvi-ously, the identity $A = I$ minimizing that term constitutes a trivial solution that is ineffective when $y$ is high dimensional. The second and third terms in (3) promote a "small" $A$ and thus prevent the estimated model to be the trivial minimizer $I$ of the first term.

Much like group lasso, the second term in (3) enforces joint group sparsity across the rows of $A$. To see this note that $\|A\|_{1,2}$ is the $L_1$ norm of the $L_2$ norms of the individual rows. Due to the sparsity promoting nature of the $L_1$ minimizer, $\hat{A}$ will have only a few rows that are not identically zero. The resulting effect is the selection of landmark dimensions $y_L$ where $L$ corresponds to the non-zero rows. We thus have that the first two terms in (3) simultaneously select the landmark dimensions $L$, and model $y_L \mapsto y$. The third term $\|A\|_1$ promotes sparsity within the coefficients of the model $y_L \mapsto y$. This additional sparsity assumption reduces the prediction risk when $y$ is high dimensional.

The regularization parameter $\lambda_1$ controls the number of landmark output dimensions. The regularization parameter $\lambda_2$ controls the sparsity of the model $y_L \mapsto y$. Both $\lambda_1$ and $\lambda_2$ should increase with $k$. When the landmark assumption holds and there exists a landmark set $L^*$ such that $y$ is a noisy sparse linear combination of $y_{L^*}$, the row sparsity pattern of $\hat{A}$ should coincide with $L^*$ (assuming an appropriate selection of $\lambda_1, \lambda_2$). As $\lambda_1/\lambda_2$ increases, the group sparsity constraints become dominant implying that each dimension of $y$ depends on all of the dimensions of $y_L$. As $\lambda_1/\lambda_2$ decreases, $\hat{A}$ tends to be more sparse within groups, implying that the dimensions of $y$ are sparse linear combinations of the $y_L$.

From a practical point of view, with a proper selection of the regularization parameters $\lambda_1, \lambda_2$ (for example using cross-validation), the model (3) is quite flexible. It allows handling situations involving large landmark sets $L$ and small landmark sets $L$, and high or low sparsity for the model $y_L \mapsto y$. Empirically, the dependence on the precise value of $\lambda_1, \lambda_2$ is robust, as small variations in $\lambda_1, \lambda_2$ do not substantially change the predicted values.

**Handling non-linear output relationship:** In order to select and learn non-linear relationships between the outputs and the landmarks, one could use functional joint sparsity models with $Ł_1/L_\infty$ constraints as proposed by (Liu et al., 2008) or with $L_1 + Ł_1/L_2$ constraints (appropriately defined on a function space). With this change in step 1, the proposed approach could be used to handle non-linear relationships between the outputs, making the proposed method more flexible. Developing concrete algorithms and analysis for this setting is left as future work.

**Step 2: Estimating** (1)

Once the landmark outputs $L$ are identified, we can proceed with fitting model (1). In the case of continuous $y$ (regression), model (1) can be estimated using a using multivariate regression model. In the case of a discrete $y$ (classification), a one vs. all classifier may be used for $x \mapsto y_i$, $i = 1, \ldots, s$, or alternatively a multiple output classifier may be used for $x \mapsto y_L$. Examples include support vector machines and log-linear models. From a statistical perspective, when $y$ is high dimensional the reduction in the number of estimated parameters from $kd$ to $sd$ (in the regression setting) where $s \ll k$, contributes to lower prediction risk. If the dimensionality of $x$ is also high, the models $x \mapsto y_L$ or $x \mapsto y_i$ should use careful feature selection or regularization to avoid overfitting.

**Step 3: Prediction**

In many cases, a statistical model for (1) provides not only point estimates, but also a full probabilistic model $p(y_L|x)$. Similarly, a statistical model for (2) provides a full probabilistic model for $p(y|y_L)$. The implied model

$$p(y|x) = p(y|y_L)p(y_L|x)$$

suggests the following procedure for predicting $y$ from $x$

$$y_L^* = \arg\max_{y_L} p(y_L|x) \tag{4}$$

$$y_{L^c}^* = \arg\max_{y_{L^c}} \int p(y|y_L)p(y_L|x)\, dy_L. \tag{5}$$

An alternative to (5) is to use the following approximation

$$\arg\max_y p(y|x) \approx \arg\max_y p\left(y|\arg\max_{y_L} p(y_L|x)\right).$$

In other words, given a new test sample $x$, we predict $y_L$ using the model from step 2, and then estimate $y_{L^c}$ using the model from step 1, operating on the predicted $y_L$. In the case of classification, we follow standard practice and set the components of $y$ to 1 if the corresponding prediction of model (2) is greater than 0.5 and to 0 if it lesser than 0.5. Finally the outputs are concatenated and they represent the prediction for the given sample $x$. Algorithm 1 summarizes this procedure.

## 4. Theory

In this section, we give a brief theoretical analysis of the proposed approach in the regression setting highlighting the advantage of the proposed approach. We assume that there exist a true landmark subset $L^*$ and provide conditions under which it could be recovered consistently. Specifically, following the analysis

**Algorithm 1** Landmark selection method

**Input:** data $\{(x_1, y_1), \ldots, (x_n, y_n)\}$ in the form of $X \in \mathbb{R}^{n\times d}$ and $Y \in \mathbb{R}^{n\times k}$
**Step 1:** Simultaneously find the landmark set $L$ and solve the optimization problem in step 1 to obtain the model $y_L \to y$ and estimate $\hat{A}$.
**Step 2:** Estimate the model $x \to y_L$ using independent models for each component of $y_L$ or using multiple-output classification or regression algorithms.
**Step 3:** Given a new test point $x$, estimate $y$ by (4)-(5).

developed in (Obozinski et al., 2011) for random design linear regression with group Lasso regularization, we can get a lower bound on the number of samples needed for recovering the support of the subset $L^*$ of the landmark labels. For simplicity, we consider the regression setting with the assumption that $\lambda_2 = 0$.

We assume that $Y$ consists of i.i.d. rows sampled from $N(0, \Sigma)$. This distribution could in fact be any sub-Gaussian distribution (which includes any bounded random variable for example the Bernoulli random variable) for which a similar analysis could be carried out. We make the following assumption on the the covariance matrix $\Sigma$: (1) there exists $\rho_{min} > 0$ and $\rho_{max} < \infty$ such that all the eigenvalues of the $s \times s$ covariance matrix $\Sigma_s$ of the the landmark output $y_L \in \mathbb{R}^s$ are contained in the closed interval $[\rho_{min}, \rho_{max}]$, (2) *mutual incoherence:* there exist a incoherence parameter $\gamma \in (0, 1]$ such that $\|\Sigma_{S^cS^c}(\Sigma_{SS})^{-1}\|_\infty \le 1-\gamma$ and (3) *self-incoherence:* there exists $D_{max} < \infty$ such that $\|(\Sigma_{SS})^{-1}\|_\infty \le D_{max}$. Note that these are standard conditions assumed for support recovery results in the modern sparse recovery analysis. Condition (1) is needed to prevent over-dependency between the landmark outputs. Conditions (2) and (3) are necessary conditions for model selection consistency of sparse recovery problems. For example, several classes of matrices, for example Toeplitz matrices, tree-structured matrices and bounded off-diagonal matrices are shown to satisfy the above conditions (Zhao & Yu, 2006). In the absence of these conditions, landmark recovery might fail even with arbitrarily large training set.

We also make the following assumption on the regression matrix. Let $a_{min} \stackrel{\text{def}}{=} \min_{i\in L} \|A_i\|_2$ where $A_i$ denote the $i^{th}$ non-zero row of the matrix $A$. We denote $A^s \in \mathbb{R}^{s\times k}$ to be the subset of the matrix $A$ with non-zero rows, $\zeta(A_s) \in \mathbb{R}^{s\times k}$ to be the row normalized matrix, and

$$\phi(A) \stackrel{\text{def}}{=} \lambda_{max}\big(\zeta(A_s)^\top(\Sigma_{SS})^{-1}\zeta(A_s)\big).$$

This quantity characterizes the amount of overlap that could be captured given the output samples. Note that the support overlap function $\phi(A)$ satisfies

$$\frac{s}{\rho_{max}K} \le \phi(A) \le \frac{s}{\rho_{min}}$$

for any $Y$ that satisfies assumption (1).

**Proposition 1.** *Consider the label matrix $Y$ with rows i.i.d. drawn from $N(0, \Sigma)$ satisfying assumptions (1)-(3), suppose that $a^2_{min}$ decays no more slowly than $f(k)\min\{\frac{1}{s}, \frac{1}{\log(k-s)}\}$ for some function $f(k)$ such that $f(k)/s \to 0$ and $f(k) \to \infty$. Then, as long as $n > C'\rho_{max}\phi(A^*)\log(k-s)$, we have with probability greater than $1 - c_1\exp(c_2\log s)$: (1) the optimization problem in 3 (with $\lambda_2 = 0$) has a unique solution when $\lambda_1 = \sqrt{\frac{f(k)\log k}{n}}$ and (2) the row support specified by the unique solution of the optimization problem 3 is equal to the row support of the true model.*

*Proof.* The proof follows from the corresponding proof in (Obozinski et al., 2011). □

The main consequence of the above proposition is that if there exist a set of landmark variable $L^*$ in the output space, the sample complexity is of *logarithmic order* in the original dimension of the output space $k$. Using sub-Gaussian assumptions on the label matrix, analogous conditions for classification are possible.

Following (Reinsel & Velu, 1998) we note that for a matrix regression problem $y = \Theta x + \epsilon$ with $\Theta \in \mathbb{R}^{m_1\times m_2}$, the Frobenious norm error rate (with $n$ samples, unit noise variance and no assumption on the regression matrix)

$$\|\hat{\Theta} - \Theta\|_F^2 = O\left(\frac{m_1 m_2}{n}\right).$$

Since in our case the estimated matrix (1) (assuming linear regression model) is of the dimension $s \times d$, the error is of the order of $O(\frac{sd}{n})$ samples (Reinsel & Velu, 1998), much smaller than the classical setting without the landmark selection method of $O(\frac{kd}{n})$. In particular, when $s \ll k$, there is a significant gain in efficiency.

We conclude that the landmark method makes a structural assumption on the output space in order to facilitate regression in high dimensional setting ($n \ll kd$). Other methods, making a different set of structural assumptions (e.g., low-rank regression) try to achieve the same goal, but work under a different set of assumptions. Empirically, the landmark method works better than low-rank regression and group Lasso based multivariate regression on a variety of datasets (see Section 6).

## 5. Optimization procedure

Here, we provide the optimization procedure required to solve the optimization problem described in step 1. The spaRSA method, proposed recently in (Wright et al., 2009), is a solver for optimization problems of the form

$$\min_{a \in \mathbb{R}^p} f(a) + \lambda \phi(a)$$

where $f$ is a convex loss function and $\phi$ is a convex regularizer. The main advantage of spaRSA is that when the regularizer is group separable, the problem decomposes over the group.

Using vectorization and block-diagonalization, it can be shown that (3) falls under this framework. Upon initial investigation, it appears that the block-diagonalization operation complicates the solver as it increases the size of the data matrix. However, we describe below a variation on spaRSA that works directly with the $Y$ and $A$. A similar approach was used in (Sprechmann et al., 2011) for the problem of collaborative dictionary learning with hierarchical penalty. The main advantage of the spaRSA procedure (that the problem decouples across groups) is still preserved and further in our case, each subproblem could be solved via thresholding.

In order to solve the optimization problem, the spaRSA procedure generates a sequence of updates that converges to the solution. We refer the reader to (Wright et al., 2009) for a complete description of the general procedure. In our case, we let $f(A_i)$ denote the reconstruction error (the squared loss in our case) for $A_i$ (here and below we denote the $i$-column of a matrix $A$ as $A_i$) and define the matrix $U^{(t)} \in \mathbb{R}^{k \times k}$ whose $i$-column is given by

$$U_i^{(t)} = A_i^{(t)} - (1/\alpha^t) \nabla f(A_i^{(t)}).$$

The sequence of spaRSA updates that converge to the true solution is

$$A^{(t+1)} = \arg\min_{Z \in \mathbb{R}^{k \times k}} \|Z - U^{(t)}\|_F + \frac{\lambda_1}{\alpha^{(t)}} \|Z\|_F + \frac{\lambda_2}{\alpha^{(t)}} \|Z\|_1,$$

which is group separable into $k$ independent problems as below:

$$A_i^{(t+1)} = \arg\min_{Z_i \in \mathbb{R}^k} \|Z_i - U_i^{(t)}\|_2 + \frac{\lambda_1}{\alpha^{(t)}} \|Z_i\|_2 + \frac{\lambda_2}{\alpha^{(t)}} \|Z\|_1.$$

The solutions for each of these sub-problems are available in closed form (similar to (Sprechmann et al., 2011)) as follows:

$$A_i^{(t+1)} = \begin{cases} \max\{0, \|h\|_2 - \lambda_1\} h / \|h\|_2 & \text{if } \|h\|_2 > 0 \\ 0 & \text{if } \|h\|_2 = 0 \end{cases}$$

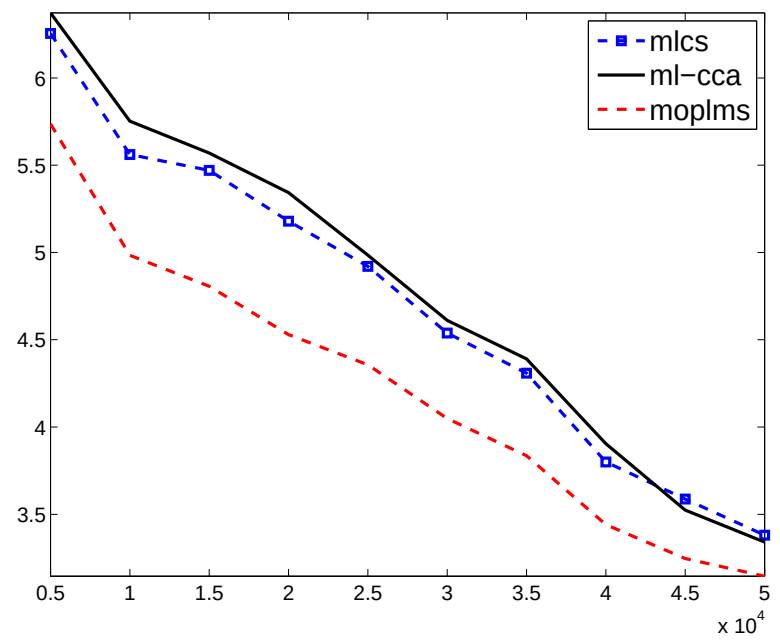

*Figure 2.* Right: Hamming loss vs. sample size on synthetic classification data sets.

where $h_j = \text{sign}(U_{i,j}^{(t)}) \max\{0, |U_{i,j}^{(t)}| - \lambda_2\}$. The thresholding require operations that are linear in the dimensionality of the matrix $Y$. The above procedure is repeated until convergence to obtain the final solution that features row sparsity, and potential sparsity within rows as well.

## 6. Experiments

In this section, we compare our landmark selection method, which we refer to below as moplms, to alternative baselines on classification and regression problems. In our experiments we used code from (Wright et al., 2009) for performing the mixed norm penalty (group lasso and lasso) landmark selection. The regularization parameters were set by cross-validation.

### 6.1. Synthetic experiments

We conducted an experiment on synthetic regression data with $k = 500$ (dimensionality of $y$), $d = 500$ (dimensionality of $x$). The number of landmark outputs $s$ was varied in the set $\{50, 100, 200\}$. The data was simulated from the above model, including the specified landmark outputs. Figure 1 (left) shows the plot of the test MSE prediction error as a function of the sample size for various values of the parameter $s/k$.

From section 4, we have that if the landmark output selection method is not used, with a linear regression model for $x \to y$,the Frobenious norm error between the true and estimated matrix scales as $O\left(\frac{kd}{n}\right)$. Where as with the landmark output assumption the error for model 1 scales as $O\left(\frac{sd}{n}\right)$. This benefit in the estimation error of the regression matrix is reflected in the MSE prediction error. Specifically, as $s$ decreases, the sample complexity decreases. This phenomenon is especially important in high-dimensional cases, when there are fewer samples than the number of parameters to be estimated. We also compared the proposed approach to group-Lasso and low-rank multivariate regression.

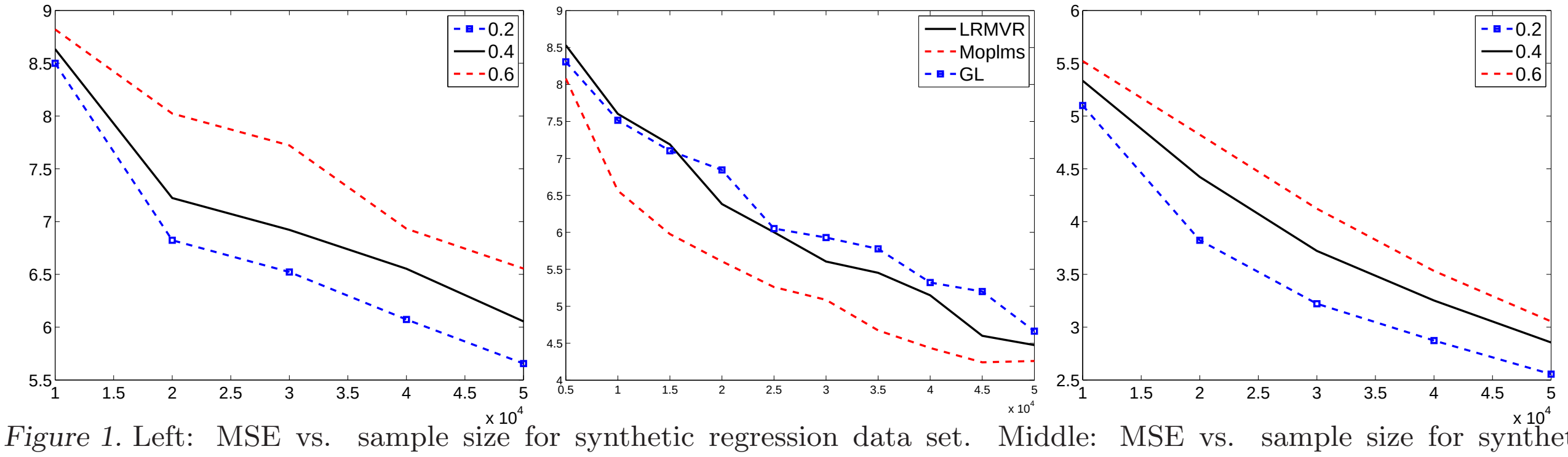

*Figure 1.* Left: MSE vs. sample size for synthetic regression data set. Middle: MSE vs. sample size for synthetic regression data set. Right: Hamming loss as a function of sample size for synthetic classification data set. The multiple curves represent different values of $s/k$.

Figure 1 (middle) shows the mse prediction error rate of the moplms method decays faster compared to the other methods.

We also experimented with synthetic classification data where $y$ is a 500 dimensional binary vector and the input $x \in \mathbb{R}^{500}$. Similar to the regression setting, the landmark outputs were first generated with $s \in \{50, 100, 200\}$ and the dependent outputs where generated as sparse linear combination of the landmark outputs. Figure 1 (right) shows the Hamming loss as a function of the sample size. The $x \mapsto y_L$ model was collection of multiple one-vs-all SVMs. Similar to the regression case, the prediction error decays with the number of landmark outputs $s/k$. We further compare the proposed approach on synthetic data set against the following methods:

1. **One vs. all:** This is a standard baseline approach for multi-label classification, for e.g., (Rifkin & Klautau, 2004).
2. **Multilabel compressive sensing (mlcs):** This approach was proposed in (Hsu et al., 2009) where the label vector is projected to a random $m$ dimensional sub-space followed by regression on the compressed subspace.
3. **Multi-label classification via canonical correlation analysis (ml-cca):** After performing canonical correlation analysis (CCA) on the input and output variables, a model is learned in the resulting subspace, followed by projection to the original label space.

From Figure 2, we note that the proposed approach has a better rate of decay of hamming loss compared to the other approaches. This phenomenon is further observed in the real world data sets as described in the next section.

We conducted an additional experiment to study the number of sub-problems selected. Specifically, we varied the number of sub-problems and the tuning parameters of mlcs, and noted the values achieving the lowest prediction error. We then trained moplms, gradually reducing the regularization parameter until the prediction error matched that of mlcs. The two methods achieved identical prediction error with the following (mean) values of $s/k$: 0.45 (mlcs) and 0.30 (moplms), indicating moplms selected fewer sub-problems while achieving identical performance. Note, however, that mlcs always uses base regressors and moplms uses base classifiers.

## 6.2. Real-world data sets

### 6.2.1. Classification

We experimented with the following two multiple output classification datasets.

1. **del.icio.us** This dataset consists of data from del.icio.us, a social bookmarking site where webpages are labeled with multiple contextual tags. The data set contains about 16000 labeled web page and 983 unique labels. We follow the experimental setup followed in (Hsu et al., 2009) and represent web page as a boolean bag-of-words vector, with the vocabulary chosen using a combination of frequency thresholding and $\chi^2$ feature ranking, resulting in 500 features.
2. **Image data set.** This dataset contains 68000 images, with about 22000 unique word tags for each image. Following (Hsu et al., 2009) we retained the 1000 most frequent labels. We represented each image via codes computed with a learned dictionary (of size 1024) via sparse coding (Yang et al., 2009). Specifically, we densely sampled $10 \times 10$ patches from the image and computed sparse codes. Finally max-pooling was used to pool the codes obtained for the patches.

Note that we use thresholding to convert the real output to the binary form of the data. The regulariza-tion parameters $\lambda_1$ and $\lambda_2$ were estimated using cross-validation. The number of selected landmarks $s$ was 231 for the del.ic.ious data and 278 for the image data set. This was less than the number of sub-problems in both the mlcs and ml-cca approaches, which were also tuned for optimal prediction error.

| | Delicious | | Image | |
|---|---|---|---|---|
| | Ham. loss | F-score | Ham. loss | F-score |
| mlcs | 0.0187 | 0.3732 | 0.0047 | 0.3012 |
| ml-cca | 0.0164 | 0.3822 | 0.0041 | 0.3183 |
| one.vs.all | 0.0144 | 0.4512 | 0.0034 | 0.3923 |
| moplms | 0.0142 | 0.4522 | 0.0032 | 0.4031 |

*Table 1.* Test set Hamming loss and F1 measure evaluation of the four classification approaches: mlcs, ml-cca, one vs. all, and moplms. The base classifiers in the reduced space were SVM.

Table 1 displays the F1-score and hamming loss that are two standard evaluation metrics for multi-label classification.

$$\text{Hamming loss} = \frac{y^\top \mathbf{1} + \hat{y}^\top \mathbf{1} - 2y^\top \hat{y}}{k}$$
$$\text{F1 score} = \frac{2y^\top \hat{y}}{\sum_{i=1}^{k} y_i + \sum_{i=1} \hat{y}_i}.$$

The landmark selection method performed better in terms of both evaluation metrics. The one-versus-all method was the second best in terms of prediction accuracy, but takes a significantly greater amount of train and test time, compared to the alternative methods.

Figure 3 (left and middle) shows the decay of the Hamming loss as a function of the sample size for mlcs, moplms and ml-cca method. We omitted the one-vs-all method as it took significantly more amount of time compared to the other approaches, and thus is not computationally attractive. The proposed landmark selection approach has lower prediction error than mlcs and ml-cca.

#### 6.2.2. Regression

In the regression setting, we consider predicting the stock prices of several companies based on previous values via the landmark selection approach on the SP 500 data set. More specifically, the data consists of closing stock prices of the 500 companies in the S&P index in the period from August 21, 2009 to August 20, 2010 (a total of 245 entries). We assume the following autoregressive 1 or AR(1) model

$$y_{t_L} = By_{t-1_L} + E \qquad (6)$$

| Method | Moplms | Group lasso | LRMV Reg |
|---|---|---|---|
| Test err | 3.28 | 5.42 | 4.63 |

*Table 2.* Test prediction error (MSE) for moplms vs. Lrmvr. $\lambda_1$ and $\lambda_2$ in state 1 were selected to minimize prediction error using cross-validation. The number of subproblems selected in this case was 98.

where $y_t = \log \frac{S_t}{S_{t-1}}$ represents the log returns ($S_t$ is the stock price at time $t$) for day $t$ and $E$ is the noise matrix. The problem is motivated by the observation in finance that multiple companies have stock prices that share identical stochastic trends (cointegration).

We compare our landmark selection approach to low-rank multivariate regression (using trace norm regularization) and group lasso based multivariate regression. These two baselines are popular multivariate regression methods. In our case (moplms), we used a multivariate ridge regression for estimating model (1), which is Equation 6 in the current setting. As in the classification setting, the regularization parameter was tuned by cross validation, and resulted in $s = 98$ landmark outputs.

Table 2 shows that moplms outperformed the two baselines (group lasso and low-rank multivariate regression). Figure 3 displays the prediction error rate as a function of the sample size. It confirms this conclusion as the prediction error of moplms decays faster than the baselines.

## 7. Discussion

In this paper we propose a framework for multi-output prediction based on parsimonious modeling on the output space. By selecting a subset of the output dimensions (landmarks) and focusing on modeling the dependency of that subset of $y$ on $x$, we reduce the sample complexity considerably. This is most noticeable when the output dimensionality is high and the different component feature high correlation. Our experiments indicate that the proposed method outperforms standard multi-output methods in both the classification and regression scenarios.

The results in this paper raise several interesting questions and follow up directions. First, a detailed analysis is required to characterize the improvements of the proposed methods over competing methods. Second it is interesting to consider cases in which the label vector has a pattern of missingness.

**Acknowledgments:** The authors would like to thank Kai Yu and Parikshit Ram for discussions.

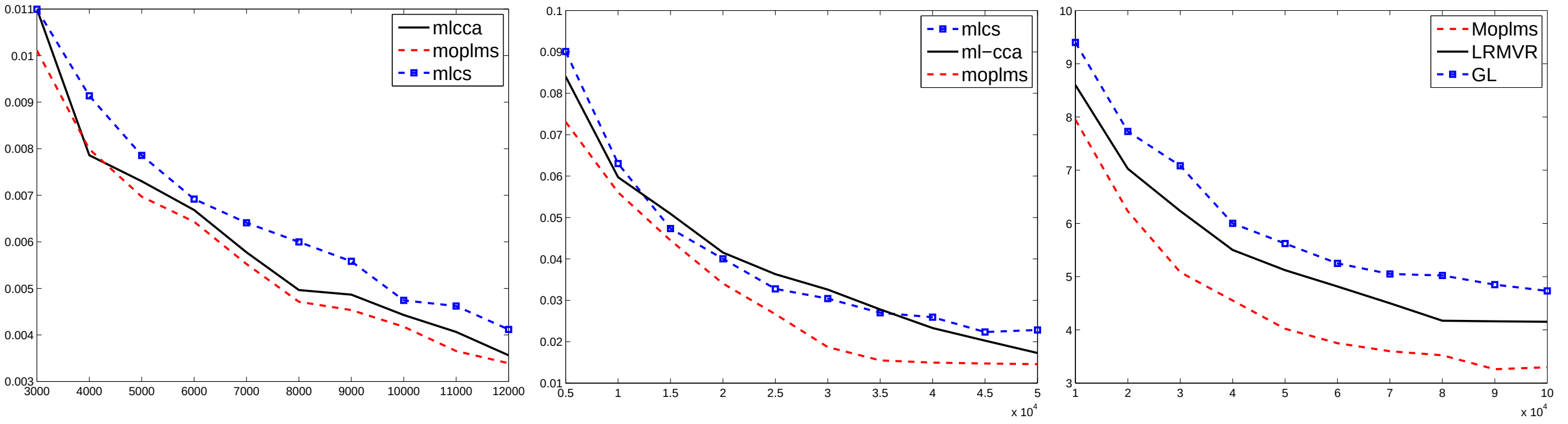

*Figure 3.* Left and middle: Hamming loss versus number of samples for moplms, mlcs and ml-cca on delicious data set (left) and image data set (middle). Right: Mean MSE prediction error as a function of sample size for moplms, low rank multivariate regression and group Lasso based multivariate regression.